\documentclass{article}

\usepackage{arxiv}

\usepackage[utf8]{inputenc} 
\usepackage[T1]{fontenc}    
\usepackage{hyperref}       
\usepackage{url}            
\usepackage{booktabs}       
\usepackage{amsfonts}       
\usepackage{nicefrac}       
\usepackage{microtype}      
\usepackage{lipsum}
\usepackage[linesnumbered,boxed,ruled]{algorithm2e}
\usepackage[noend]{algpseudocode}
\usepackage{multirow}
\usepackage{enumitem}
\usepackage{graphicx, subfigure}
\usepackage{mathtools}
\usepackage{color}
\DeclareMathOperator*{\argmax}{arg\,max}

\newcommand{\nosection}[1]{\vspace{1.5pt}\noindent\textbf{#1.}}
\title{ASFGNN: Automated Separated-Federated Graph Neural Network}

\author{
  Longfei Zheng \\
  Ant Group, Beijing, China\\
  \texttt{zlf206411@antgroup.com} \\
   \And
  Jun Zhou \\
  Ant Group, Beijing, China\\
  \texttt{jun.zhoujun@antgroup.com} \\
  \And
  Chaochao Chen\thanks{Corresponding author.} \\
  Ant Group, Hangzhou, China\\
  \texttt{chaochao.ccc@antgroup.com} \\
  \And
  Bingzhe Wu \\
  Ant Group, Beijing, China\\
  \texttt{fengyuan.wbz@antgroup.com} \\
  \And
  Li Wang \\
  Ant Group, Hangzhou, China\\
  \texttt{raymond.wangl@antgroup.com} \\
  \And
  Benyu Zhang \\
  Ant Group, Sunnyvale, US\\
  \texttt{benyu.z@antgroup.com} \\
}

\begin{document}
\maketitle

\begin{abstract}
Graph Neural Networks~(GNNs) have achieved remarkable performance by taking advantage of graph data.
The success of GNN models always depends on rich features and adjacent relationships.
However, in practice, such data are usually isolated by different data owners (clients) and thus are likely to be 
Non-Independent and Identically Distributed (Non-IID). 
Meanwhile, considering the limited network status of data owners, hyper-parameters optimization for collaborative learning approaches is time-consuming in data isolation scenarios.
To address these problems, we propose an Automated Separated-Federated Graph Neural Network~(ASFGNN) learning paradigm. ASFGNN consists of two main components, i.e., the training of GNN and the tuning of hyper-parameters. 
Specifically, to solve the data Non-IID problem, we first propose a separated-federated GNN learning model, which decouples the training of GNN into two parts: the message passing part that is done by clients separately, and the loss computing part that is learnt by clients federally. 
To handle the time-consuming parameter tuning problem, we leverage Bayesian optimization technique to automatically tune the hyper-parameters of all the clients. 
We conduct experiments on benchmark datasets and the results demonstrate that ASFGNN significantly outperforms the naive federated GNN, in terms of both accuracy and parameter-tuning efficiency. 
\end{abstract}

\keywords{Graph neural network \and Federated learning \and Bayesian optimization \and Privacy preserving}

\section{Introduction}
\label{intro}
Graph Neural Networks~(GNNs) have achieved superior performance by taking advantage of embedding features via aggregating representations of nodes and their neighbors \cite{Scarselli2009GNN}. GNNs benefit a lot of applications across different tasks, such as computer vision \cite{wang2018dynamic}, traffic prediction \cite{Thomas2018GCN}, recommend system \cite{ying2018graph} and risk control \cite{liu2018}.
\subsection{Existing problem}
\label{problem}
The factor that drives the success of GNN is the rapid growth of high-dimensional data and their adjacent information. 
However, existing GNN methods face two main challenges. 
First of all, with the increasing awareness of security and privacy, \emph{data-isolation} problem is serious, which limits the data size of a single party (client) and further damage the performance of GNN. 
Furthermore, the isolated datasets in different clients are usually Non-Independent and Identically Distributed (Non-IID), due to the reasons that clients belong to diverse geographic locations or have different time windows of data collection. 
Therefore, it becomes more and more difficult to train a global GNN model with the Non-IID data in data isolation scenario.

\begin{figure}[ht]
  \centering
  \includegraphics[width=0.75\textwidth]{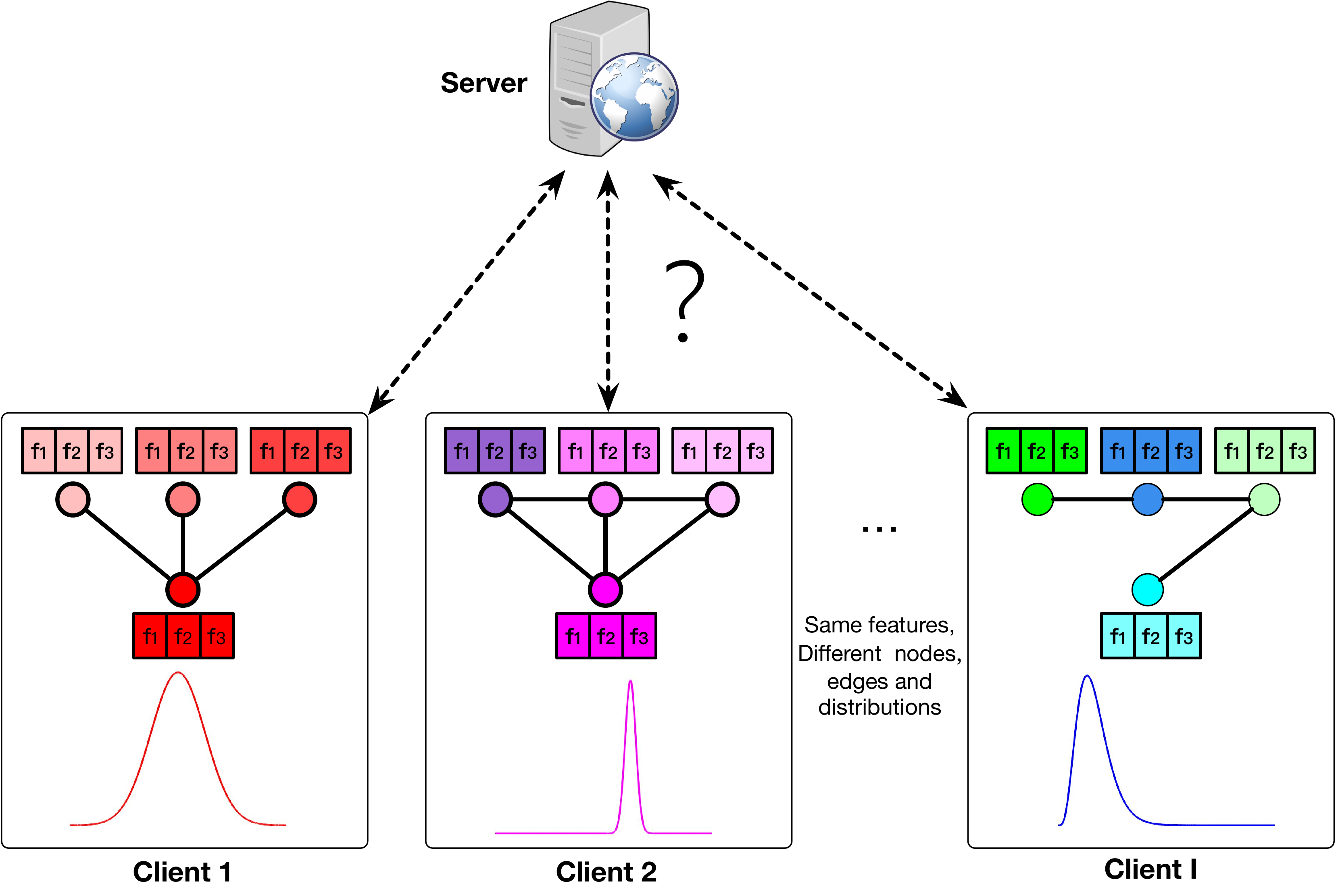}
  \caption{ The data isolation problem with Non-IID graph data, assuming $I$
  clients with four nodes, three features and different label distributions.}
  \label{Isolated problem}
\end{figure}

Fig. \ref{Isolated problem} shows a typical example of the Non-IID graph data, where we assume there are $I$ separated clients. 
These clients collect graph data from different sources with the same format. 
In other words, clients share the same feature domain, e.g., $ \{ f_1$,$f_2$,$f_3 \}$, but differ in sample space, which are represented by colorful nodes. 
Meanwhile, clients may have diverse graph structures of nodes, i.e., heterogeneous graphs. Furthermore, data distributions are likely to be Non-IID , as is shown in Fig. \ref{Isolated problem}. 

Moreover, hyper-parameters are important for GNN learning algorithms. For example, activation function determines the output of layers, regularization parameter impacts the calculation of loss functions, and learning rate influences the update of model weights in the back-propagation process \cite{Lorenzo2017hyper}.
These hyper-parameters directly affect the training process of GNN models. 
Intuitively, in order to achieve the best model performance, clients with Non-IID graph data are likely to have individual hyper-parameter sets rather than a global hyper-parameter set \cite{kairouz2019advances}. 
Due to the huge search space and limited network status among clients, tuning of hyper-parameters is quite time-consuming. 
Therefore, it is important to design a proper distributed GNN model on Non-IID dataset with hyper-parameter optimization power.

Unfortunately, there is few literature on solving the above problem. 
Although directly applying federated learning to GNN seems a good choice, it has two main shortcomings \cite{kairouz2019advances}.
Firstly, federated learning faces the statistical challenge. 
The original goal of federated learning, i.e., training a single global model on the union of clients' datasets, is no longer suitable for Non-IID graph data \cite{zhao2018federated}. 
Secondly, communication of federated GNN learning is time-consuming. 
This is because, in order to achieve the best performance, models and hyper-parameters of clients are likely to be different. 
Comparing with the traditional neural network, GNN has extra individual hyper-parameters to represent graph information, which further increases the unbearable training time.

\subsection{Our Solution}
\label{solution}
In order to bridge these gaps, we propose an Auto Separated-Federated GNN~(ASFGNN) learning paradigm. 
As graph data is often owned by companies and governments, we focus on the cross-silo federated learning in which the clients are a limited number of organizations with powerful computing ability and reliable communications \cite{kairouz2019advances}.
Our proposed ASFGNN consists of two steps, i.e., GNN training and hyper-parameters optimization. 

In the first step, the Separated-Federated GNN learning framework decouples a GNN model into two parts: \textit{message passing} sub-model that is conducted by clients separately and \textit{loss computing} sub-model which is performed by clients federally. 
Specifically, clients first perform message passing, i.e., neighbor information aggregation, individually, and get node embeddings.
In the following step, clients take the embeddings as the input of the discrimination model to compute loss, then update both message passing sub-model and loss computing sub-model using backward propagation for the first time.
After it, the server securely aggregates the local discrimination models using federated learning method and gets the global discrimination model.
Finally, the global discrimination model is broadcast to clients to update the local discrimination models with the help of Jensen–Shannon divergence. 

In the second step, we propose a \textit{Bayesian optimization algorithm} to automatically optimize the hyper-parameters of Separated-Federated GNN model. 
That is, Bayesian optimization algorithm takes hyper-parameters as input and regards the average value of clients' evaluation metrics (e.g., precision) as output \cite{chen2019techniques,yu2020hyper}, where these metrics are uploaded by clients and averaged by server in a secure manner. 
To this end, we get the hyper-parameters that achieve the best metric.

To verify the performance of our proposed ASFGNN, we empirically compare the accuracy of SFGNN and traditional federated GNN model, and analyze the efficiency of Bayesian optimization method and the traditional grid search method.

We summarize our main contributions as follows:
\begin{itemize}
\item We propose a novel Separated-Federated Graph Neural Network (SFGNN) learning framework, which can be used to learn any existing GNN models under privacy consideration.
\item We propose to adopt Bayesian optimization to tune model parameters automatically, which significantly improves the efficiency of the SFGNN model.
\item We conduct experiments on three benchmark datasets and the results demonstrate that our proposed SFGNN outperforms federated GNN in terms of accuracy, and ASFGNN significantly reduces the hyper-parameter tuning time of SFGNN comparing with grid search.
\end{itemize}
\section{Related work}
\label{rw}
In this section, we briefly review the literature on federated learning and hyper-parameters optimization. 
\subsection{Federated learning}
\label{fed}
Federated learning model is prevailing privacy-preserving approach via model or gradient aggregation rather than data aggregation \cite{McMahan2016FederatedLO}. 
However, the accuracy of federated learning would drop significantly with Non-IID datasets\cite{kairouz2019advances}. 
Existing works propose different strategies to resolve the statistical challenge of federated learning. 
One natural approach is to create a small shared dataset which makes the data across clients more similar \cite{wang2018dataset}. 
For some applications, the contributions of clients to the global model are bounded according to the dataset characteristics \cite{thakkar2019differentially}.
Furthermore, model-agnostic meta-learning has been developed to meta-learn a global model, which can be used as a starting point for learning a good model of Non-IID data in each client \cite{finn2017modelagnostic}. 
These methods modify federated learning model with Non-IID datasets, which can not be applied in GNN model directly.
As GNN model includes two parts as shown in preliminary, among which the message passing part owns personal information which should be learned individually.

Besides the federated learning, Split Learning~(SL) is another decentralized method which trains the local models separately and sends hidden layers to server \cite{gupta2018distributed}. 
The separated local models represent the personality of clients with Non-IID datasets \cite{gao2020end}.
However, it is obviously that the hidden layers leak privacy information and the deep local layers decrease the accuracy seriously \cite{abuadbba2020can}. 
In this paper, we combine the advantages of federated learning and split learning, and propose a novel Separated-Federated Graph Neural Network learning framework.

\subsection{Hyper-parameters optimization}
\label{bo}
Recently, there has been an increasing literature on hyper-parameters optimization \cite{yu2020hyper}. 
Grid search is the most traditional way of hyper-parameters tuning, which enumerates every possible configuration in the search space. 
Random search is better than naive grid search, which samples configurations randomly.
Moreover, Evolutionary Algorithm~(EA) and Reinforcement Learning~(RL) methods are used to generate a new population (a bunch of configurations).
Another conventional solution resorts to formalizing machine learning process as a black-box optimization task, reference \cite{Swersky2013Swersky} finds the optimal of black-box objectives with the method of Bayesian Optimization~(BO).
Comparing with EA and RL, BO is more efficient than these methods and shows promising results in hyper-parameters optimization \cite{yu2020hyper}.
In this paper, we propose to apply BO as a prevailing approach to find the proper hyper-parameters in our proposed model.
\section{Preliminaries}
\label{pre}
In this section, we present some preliminary techniques and methods of our proposal, including Graph Neural Network~(GNN), federated learning, secret sharing, Jensen-Shannon divergence, and Bayesian optimization.
\subsection{Graph neural network}
GNN learns node embeddings by aggregating features of node and its neighbors.
The node embeddings are regarded as the new node representations which are fed to downstream machine learning tasks. 
The process of GNN training includes two steps: message passing and loss computing. 
The first step is the difference between GNN and other neural network models, which uses a generation function to infer node embeddings.
Numbers of message passing functions have been proposed, e.g., random walk statistics based, attention based, similarity based, and convolution based \cite{perozzi2014deepwalk,velikovi2017graph,mei2019sgnn,Thomas2018GCN}.
In this work, we select GraphSAGE as the node embeddings generation function, which aggregates the embeddings from a node’s local neighborhood in a inductive way \cite{hamilton2017inductive}.
The message passing process is described in Equation \eqref{GraphSAGE}, where $k \in \{1,2,...,K\}$ denotes the depth of neighborhood aggregation, $\textbf{h}_{v,k}$ denotes the embedding of node $v$ during $k$-th aggregation step, $\textbf{x}_v$ denotes features of node $v$ and $\mathcal{N}(v)$ denotes the neighbors of node $v$ in the graph \cite{wu2019comprehensive}. 
\begin{equation}
\label{GraphSAGE}
\begin{split}
& \textbf{h}_{\mathcal{N}(v),k} \leftarrow \text{AGG}_k(\{\textbf{h}_{u,k-1}, \forall u \in \mathcal{N}(v)\}),\\
& \textbf{h}_{v,k} \leftarrow \left(\textbf{W}_k \cdot \text{CONCAT} \left( \textbf{h}_{v,k-1},\textbf{h}_{\mathcal{N}(v),k} \right) \right).
\end{split}	
\end{equation}
where AGG$_k$ is the aggregation function in $k$-th step, such as Mean, LSTM, and Pooling methods \cite{hamilton2017inductive}. 
\subsection{Federated learning}
\label{fl}
Federated Learning (FL) was first proposed by Google \cite{McMahan2017CommunicationEfficientLO}, which builds distributed machine learning models while keeping personal data on clients.
In other words, federated learning models are trained via model aggregation rather than data aggregation.
We suppose that $I$ clients have their own datasets $\{D_1,D_2,...,D_I\}$ which are collected from different sources with the same feature domain. 
Private raw dataset $D_i$ is preserved locally, client $i$ uses forward and backward propagations to update its own model $M_i$ individually, which has the identical neural network architecture with other clients. 
Then clients upload the encrypted weights to the server with the help of secret sharing or homomorphic encryption \cite{shamir1979share,aono2016scalable,zhou2020privacy,chen2020homomorphic}. 
The server averages the uploaded model parameters to update the global federated model $M_s$, which will be sent back to client $i$ to replace the local model $M_i$.
\subsection{Jensen-Shannon divergence}
\label{js}
The Jensen–Shannon divergence~(JS) is popularly used to evaluate the dissimilarity between two probability distributions \cite{js1990anew}. 
JS has a finite value range from 0 to 1 for two probability distributions.
Motivated by \cite{Bojchevski2017DeepGE}, JS can be used to indicate the dissimilarity between two None-IID datasets. 
Considering two probability distributions $P$ and $Q$, the JS between $P$ and $Q$ is defined in Equation \eqref{JS}. 
\begin{equation}
\label{JS}
\begin{split}
&   \textbf{JS}\left (P||Q \right ) \leftarrow \frac{1}{2} \textbf{KL} \left (P|| \frac{P+Q}{2} \right ) + \frac{1}{2} \textbf{KL} \left (Q|| \frac{P+Q}{2} \right ),\\
&   \textbf{KL}\left (P_1||P_2 \right ) \leftarrow \sum_{x \in X}P_1(x)\log{\frac{P_1(x)}{P_2(x)}}.
\end{split}
\end{equation}

As the machine learning model is built to represents the trained dataset, the difference between the aggregated model in server and the local model in client can be simulated by the distribution similarity between the participated data and the client data.

\subsection{Secret sharing}
\label{ss}
 Our proposal depends on Shamir's $t$-out-of-$n$ threshold secret sharing algorithm \cite{shamir1979share}. 
 Typically, we use $n$-out-of-$n$ additive secret sharing to recover the privacy in this paper. 
 For example, we suppose that there is an $\ell$-bit value $a$ of client $i, i \in \mathcal{P}$ with $ \mathcal{P}= \{1,...,I\} $, which will be shared among all the participant clients. 
 Firstly, in order to encrypt $(\textbf{Shr}(\cdot))$ the value $a$ of client i, client $i$ generates a random number $a_j$, $\{a_j \in \mathcal{Z}_{2^\ell}, j \in \mathcal{P}, j \neq i\}$, which will be distributed to client $j, \{j \in \mathcal{P}, j \neq i\}$. 
 Then client i calculates $a_i = a- \sum_j a_j$ mod $2^\ell$ which will be kept locally.
 For simplification, We use $\langle a \rangle_k $ to denote the share of $a$ in client $k$, $ \forall k \in \mathcal{P}$. 
 To decrypt $(\textbf{Rec}(\cdot))$ the shared value $a $, client $k$ $ ( \forall k \in \mathcal{P} )$ sends the encrypted value $\langle a \rangle _k$ to the server. The server aggregates $ \sum_k \langle a \rangle _k$ mod $2^\ell, k \in \mathcal{P}$ , and gets the value $a$ of client $i$.
\subsection{Bayesian optimization}
\label{baye}
Bayesian Optimization~(BO) is an effective method to solve the black-box parameter optimization problem \cite{yu2020hyper}. In our paper, we care about the hyper-parameter optimization in the training of GNN model, where we try to find the optimal hyper-parameter setting that maximizes the utility function:
\begin{equation}
\theta^* = \argmax_{\theta\in\Theta} f(\theta),
\label{eq:black-box}
\end{equation}
where $\theta$ denotes the hyper-parameters, such as learning rate and dimension of hidden units. The $\Theta$ denotes the search space and $f$ is the utility function which is measured by certain model metrics, such as model accuracy and the Area Under Curve (AUC) score.
Typically, the evaluation of $f$ is expensive and we cannot obtain its closed-form solution.
Therefore, we treat Equation \eqref{eq:black-box} as a black-box optimization and adopt BO to solve this problem. The key ingredients of BO include a surrogate model to ``imitate'' $f$ and an acquisition function to decide the next trial based on historical trails (i.e., hyper-parameters). In our paper, we use Gaussian process~(GP) as our surrogate model and use the Expected Improvement~(EI) function as the acquisition function \cite{WU2019Hyperparameter}.  
\section{The proposed method}
\label{method}
\begin{figure*}[t]
\centering
\subfigure[Separated federated graph neural network~(SFGNN) model]{ \includegraphics[height=6.5cm]{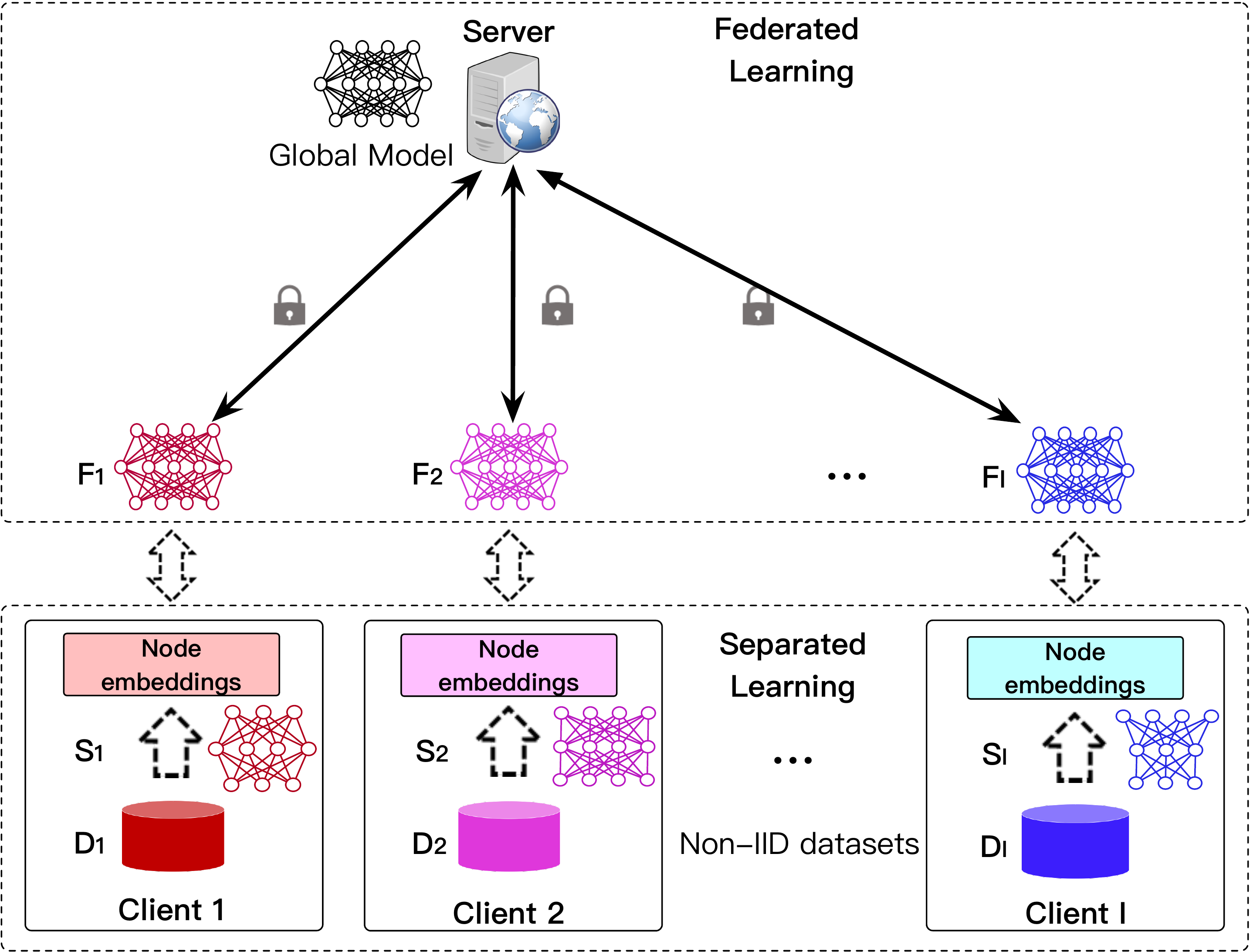}}~~~
\subfigure[Hyper-parameters Bayesian optimization of SFGNN]{ \includegraphics[height=6.5cm]{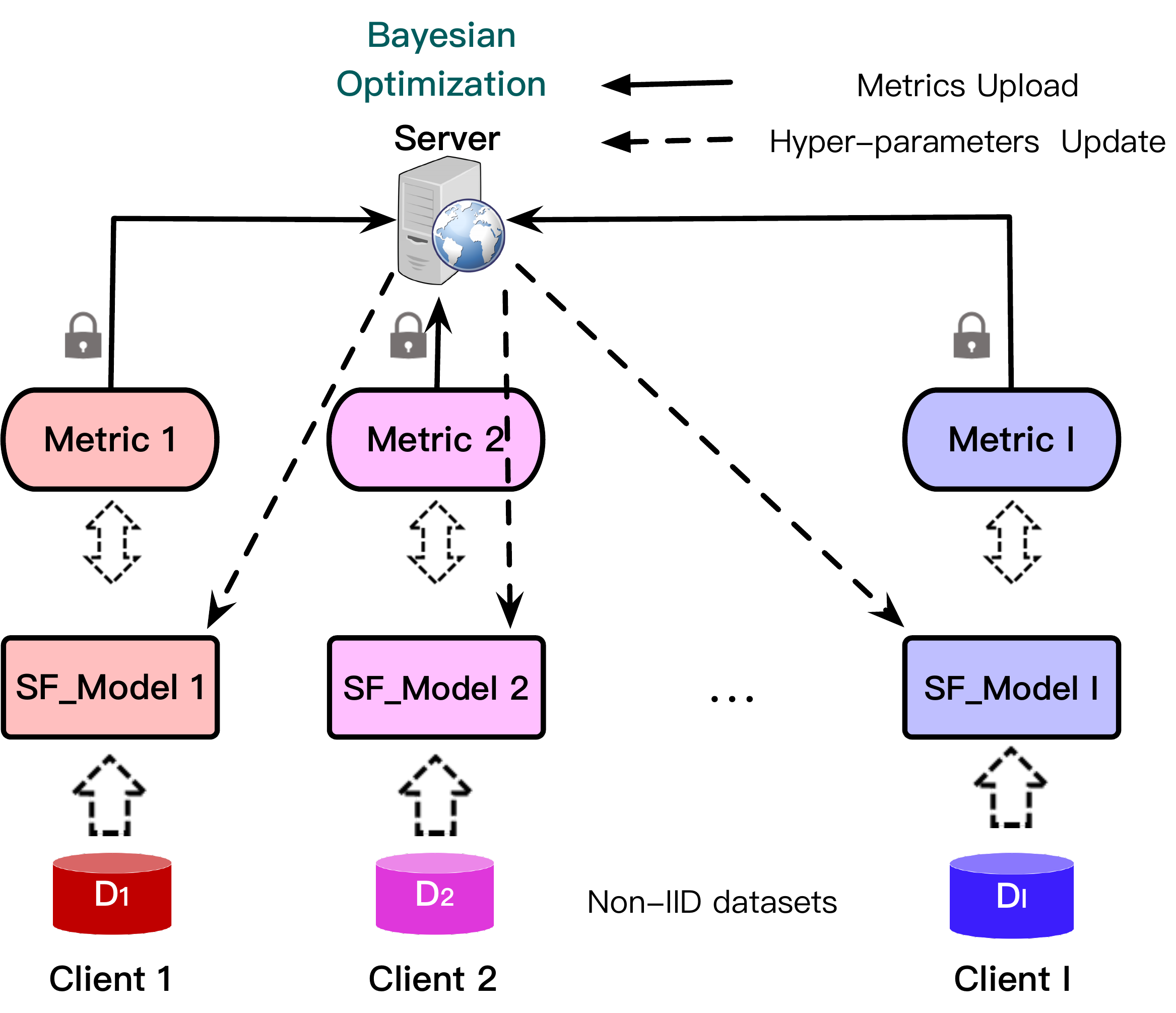}}
\caption{Our proposed automated separated-federated graph neural network model.}
\label{SFGNN}
\end{figure*}
\begin{table}[ht]
\centering
\caption{Notations and descriptions.}
\label{notations}
\begin{tabular}{c|p{5cm}|c|p{5cm}}
  \hline
  Notation & Description &Notation & Description \\
  \hline
    $I$ & total number of clients & $\mathcal{P}$ & union set of $I$ clients\\
    $G^i$ & graph data of client $i$ & $V^i$ & nodes data of client $i$ \\
    $E^i$ & edges data of client $i$ & $\mathcal{N}^i(v)$ & neighbour function of client $i$\\
    $\textbf{x}_v^i$ & features of node $v$ in client $i$ & $J$ & total number of categories\\
    $W_{k,t}^i$ & weights of $k$-th step in SGNN for client $i$ during $t$-th epoch & $K$ & depth of neighbor aggregation in SGNN\\
    $h_{v,k}^i$ & intermediate node embeddings of node $v$ in client $i$ during $k$-th step & $H_{v,t}^i$ & the final node embeddings of node $v$ in client $i$ during $t$-th epoch \\
    $Q^i_t$ & probability density of label in client $i$ during $t$-th epoch & $Q^s_t$ & probability density of label in server during $t$-th epoch\\
    $N^i_t$ & sample numbers of different categories in client $i$ during $t$-th epoch & $N^s_t$ & sample numbers of different categories in server during $t$-th epoch\\
    $n^i_{j,t}$ & sample numbers of category $j$ in client $i$ during $t$-th epoch & $n^i_{t}$ & total number of samples in client $i$ during $t$-th epoch\\
    $y_t^i$ & labels in client $i$ & $\hat{y_t}^i$ & labels prediction in client $i$\\
    $W_{l,t}^i$ & intermediate weights of $l$-th layer in FGNN of client $i$ during $t$-th epoch & $\overline W_{l,t}^i$ & weights of $l$-th layer in FGNN of client $i$ during $t$-th epoch\\
    $\overline W_{l,t}^s$ & weights of $l$-th layer in FGNN model of server during $t$-th epoch & $js^i_t$ & JS divergence between dataset of client $i$ and dataset of server during $t$-th epoch\\
    $\textbf{Shr}(\cdot)$ & additively secret sharing encrypt & $\textbf{Rec}(\cdot)$ & additively secret sharing decrypt\\
    $\langle \cdot \rangle$ & encryption using secret sharing & $L$ & number of layers in FGNN model\\
    $M_t$ & average of metrics during $t$-th epoch & $M_t^i$ & metric of client $i$ during $t$-th epoch \\
    ${lr}^i_n$ & learning rate of client $i$ in $n$-th BO round & ${l_2}^i_n$ & L2 regularization of client $i$ in $n$-th BO round\\
    $\theta_{n}$ & hyper-parameters set in $n$-th BO round & $M(\cdot)$ & black-box function of hyper-parameters optimization\\
\hline
\end{tabular}
\end{table}
In this section, we first give an overview of the proposed Automated Separated-Federated Graph Neural Network~(ASFGNN) learning paradigm. We then present its three main components, i.e., separated learning for message passing on clients, federated learning for loss computing with Jensen–Shannon divergence, and hyper-parameters optimization with Bayesian optimization. Finally, we summarize the whole algorithm. 
\subsection{Overview}
\label{overview}
We first give an overview of the proposed ASFGNN learning framework. 
We focus on horizontally split datasets in this paper.

Our design of ASFGNN consists of two steps. First, we need to design a privacy preserving GNN learning model, which can solve the Non-IID problem and reduce the communication cost as much as possible. 
Second, since GNN has many hyper-parameters, we need to design a strategy to automatically optimize hyper-parameters to reduce the training time. 

The first step is to design a practical GNN learning paradigm without leaking the private plaintext data of clients. 
Inspired by existing works \cite{gu2019securing,zheng2020industrial}, we propose a Separated-Federated GNN~(\textbf{SFGNN}) learning framework. 
The main idea is decoupling the computation module of GNN into two sub-modules, i.e., the Separated GNN learning~(\textbf{SGNN}) model and the Federated GNN learning~(\textbf{FGNN}) model, as shown in Fig. \ref{SFGNN} (a). 
The former performs message passing and obtains the node embeddings as inputs of the latter one.
As clients have Non-IID datasets, node embeddings are generated separately with individual network architecture and hyper-parameters.
After the generation of node embeddings with SGNN, FGNN trains the discrimination neural network taking advantage of federated learning algorithm.

Secondly, hyper-parameters of SFGNN, such as learning rate, regularization factor, network structures etc., explode with the increasing number of clients. We adopt Bayesian Optimization method to solve this black-box optimization problem, in which we regard the hyper-parameters of model as inputs and the average of clients' metrics as outputs, as shown in Fig. \ref{SFGNN} (b). The metrics of SFGNN model in clients are securely aggregated in server. Then the server optimizes the hyper-parameters and sends the hyper-parameters back to clients to finish another training epoch of SFGNN.
To the end, the whole parameter-tuning time is greatly decreased, as the searching round of hyper-parameters is highly reduced.

In summary, we leverage Bayesian optimization technique to automatically tune the hyper-parameters of SFGNN model, combining SGNN with FGNN.
\nosection{Notations}
\label{notation}
Before presenting our model in details, we first describe the notations. 
Considering there are many notations, for clarity, we summarize the notations used in this paper in Table \ref{notations}.
\begin{algorithm}[t]
\caption{\textbf{SGNN} (Separated GNN learning on client)}\label{Separated Learning}
\KwIn {Graph $G(V^i,E^i)$ and node features \{$\textbf{x}_v^i, \forall v \in V^i$\} on data holder $i,i \in \mathcal{P} $;
depth $K$; non-linearity function $\sigma$; neighborhood functions $\mathcal{N}^i(v): v \rightarrow 2^{V^i}, \forall i \in \mathcal{P}$}
\KwOut{Node embeddings:$H_{v,t}^i, \forall v \in V^i$ on client $i$ during $t-th$ training round}
\# Calculate the initial node embeddings\\ $h_{v,0}^i \leftarrow \textbf{x}^i_v\cdot W^i_0, \forall i \in \mathcal{P}, \forall v \in V^i$\\
\# Generate local node embeddings\\
\For{each round $t=1,2,...,T$ }
{
    \For{$i \in \mathcal{P}$ \texttt{in parallel}} 
    {
        \For{$k=1$ to $K$}
        {
            \For{$v \in V$}
            {
                \textbf{client $i$}: calculates $h_{v,k}^{\mathcal{N}^i(v)} \leftarrow$ Mean$\left ( \left \{h_{u,k-1}^i \> , \forall u \in \mathcal{N}^i(v) \right \}\right )$
            }
            \textbf{client $i$}: calculates $h_{v,k}^i \leftarrow \sigma \left (W^i_{k,t} \cdot CONCAT \left (h^i_{v,k-1} ,h_{v,k}^{\mathcal{N}^i(v)} \right ) \right)$
        }
        \textbf{Client} $i$: calculates $H_{v,t}^i\leftarrow h_{v,K}^i/ ||h_{v,K}^i||_2, \forall v \in V^i$
    }
    \Return Node embeddings $ H_{v,t}^i, \forall i \in \mathcal{P}$
}
\end{algorithm}
\subsection{Separated GNN learning (SGNN)}
\label{SGNN}
We summarize how to generate initial node embeddings for client $i ( i \in \mathcal{P})$ using GraphSAGE method \cite{hamilton2017inductive} in SGNN Algorithm \ref{Separated Learning}, where the entire graph $G^i =(V^i, E^i)$, features for all nodes $\textbf{x}_v^i$ $\left ( {\forall v \in V^i} \right )$ are provided as inputs. 
The weight matrix $W_k^i, \forall k \in \left\{ 1, ...,K \right\}$ are used to propagate information of message passing layers.
The first step is generating initial node embeddings using nodes' private features, e.g., user features in social networks (line 2).
In the next step, clients generate local node embeddings by aggregating multi-hop neighbors' information using GraphSAGE method \cite{hamilton2017inductive} for the FGNN computations as shown in line 4-15 in Algorithm \ref{Separated Learning}.

\begin{algorithm}[t]
\caption{\textbf{FGNN} (Federated GNN learning)}\label{federated_learning}
\KwIn{Node embeddings $H_{v,t}^i, \forall v \in V, \forall i \in \mathcal{P}$; hyper-parameters set$\theta_n$}
\KwOut{$M_t$}
\# Client model update \\
Randomly initialization $\overline W_{l,0}^i, \forall i \in \mathcal{P} , \forall l \in \{1,...,L\}$\\
\For{each round $t=1,2,...,T$ }
{
    \# Updates local FGNN model's weights and sends to server\\
    \For{$i \in \mathcal{P}$ \texttt{in parallel}} 
    {
        \# Get: $Q_{t}^i, N_{t}^i$ \\
        $H_{v,t}^i \leftarrow$ \textbf{SGNN} $\left( G(V^i,E^i),\mathcal{N}^i(v),\textbf{x}_v^i \right)$\\
        $\hat{y^i_t} \leftarrow \sigma \left ( H_{v,t}^i \cdot \overline W_{l,t}^i\right)$ \\
        \# Get: $L(\hat{y_t}^i,y^i_t)$, $M_{t}^i$ \\
        $W_{k,t}^i \leftarrow W_{k,t-1}^i - {lr}^i_m \nabla L \left (\hat{y_t}^i,y^i_t \right )$\\
        $W_{l,t}^i \leftarrow \overline W_{l,t-1}^i - {lr}^i_m \nabla L \left (\hat{y_t}^i,y^i_t \right )$\\
        \# Upload privacy information using secret sharing:\\
        $\langle W_{l,t}^i \rangle, \langle N^i_t\rangle, \langle M^i_t \rangle$ $\leftarrow$ $\textbf{Shr}(W_{l,t}^i), \textbf{Shr}(N_{t}^i), \textbf{Shr}(M_{t}^i) $, upload to server\\
    }
    \# Secure aggregation in server:\\
    $\overline W_{l,t}^s \leftarrow \frac{1}{I} \left ( \sum_{i=1}^I \langle W_{l,t}^i \rangle \right )$\\
    $N_{t}^s \leftarrow \sum_{i=1}^I \langle N^i_t\rangle $\\
    $Q_{t}^s \leftarrow \frac{N_{t}^s}{\sum_{J} N_{t}^s}$\\
    $M_t = \frac{1}{I}\sum_{i=1}^I \langle M^i_t\rangle$\\
    \# Send $\overline W_{l,t}^s,Q_{t}^s$ to client $i, \forall i \in \mathcal{P}$\\
    \# Update $\overline W_{l,t}^i$ in client:\\
    \For{$i \in \mathcal{P}$ \texttt{in parallel}} 
    {
        $js^i_t \leftarrow JS \left (Q_t^i||Q_t^s \right)$\\
        $\overline W_{l,t}^i \leftarrow js^i_t \cdot \overline W_{l,t}^i +\left (1-js^i_t \right ) \cdot W_{l,t}^s$\\
    }
    \Return $M_t$
}
\end{algorithm}
\subsection{Federated GNN learning (FGNN) }
\label{FGNN}
First of all, client $i \left (\forall i \in \mathcal{P} \right )$ randomly initializes weights of Federated GNN Learning model $\overline W_{l,0}^i, l \in \{1,...,L\}$ with $L$ denoting the max layer. 
Client $i$ gets the label distribution $Q^i_t$ $\left (Q^i_t= \left \{ q_{t,1}^i,q_{t,2}^i,...,q_{t,J}^i \right \}\right )$ in the current batch during training epoch $t$ with $ n^i_t$ samples, where $J$ denotes the label classification as shown in FGNN Algorithm \ref{federated_learning}. 
Then client $i$ counts sample numbers of different categories $N^i_{t} = \{ n_{t,1}^i,n^i_{t,2},...,n^i_{t,J} \}$, where $\sum_{j=1}^J n_{t,j}^i = n^i_t$ ( line 4). 
Meanwhile client $i$ updates local FGNN model's weights $W_{k,t}^i$ and ${W}_{l,t}^i$ using forward and backward propagation with their own embeddings $H_{v,t}^i$ generated by Algorithm \ref{Separated Learning} (lines 5-11).
Loss function $L(\hat{y_t}^i,y_t^i))$ is defined by different tasks, e.g., cross entropy loss for classification task and mean absolute loss for regression task. In this paper, we choose classification task for example, the loss of which is defined in Equation \eqref{Lossfunction}. 
\begin{equation}
\label{Lossfunction}
\begin{split}
   L(\hat{y_t}^i,y_t^i) = - \frac{1}{n_t^i}\sum_{j=1}^J \hat{y}_{j,t}^i\log^{y_{j,t}^i}+{ l_2}^i_n \cdot \left(\sum_{k=0}^K||W^i_{k,t}||_2 + \sum_{l=1}^L||\overline W^i_{l,t}||_2\right).
\end{split}
\end{equation} 
After it, $W_{l,t}^i$ , $N^i_t$, and $M^i_t$ $\left ( i \in \mathcal{P}, l \in \{1,...,L\}\right )$ of clients are uploaded to server with the help of secret sharing $(\textbf{Shr}(\cdot))$, supposing all clients participate in the federated learning, as shown in line 13.
The server aggregates the global FGNN model $\overline W_{l,t}^s$ by averaging the sum of $W_{l,t}^i$, and gets the global label distribution $Q^s_t$, sample numbers $N^s_t$ of a training batch and average of metrics $M_t$, all of which are regarded as outputs of FGNN model, as shown in Algorithm \ref{federated_learning} line 16-19.
Then $\overline W_{l,t}^s$ and $Q^s_t$ are sent back to clients.
To the end, client $i$ calculates $js_t^i$ with the help of $Q^i_t$ and $Q^s_t$, then the local FGNN model is updated by combining $\overline W_{l,t}^s$ and $W_{l,t}^i$ (line 24). 
$js_t^i$ controls the percent of the client local model in update process. The more Non-IID clients datasets are, the bigger priority of client model is. In a world, the addition of JS contributes to the accuracy of client model in Non-IID federated learning.
\begin{algorithm}[t]
\caption{Hyper-parameters optimization}\label{Bayesian optimization}
Place a Gaussian process prior on $f$\\
Observe $f$ at $n_0$ hyper-parameters groups according to an initial space-filling experimental design\\
Set $n = n_0$\\
\While{$n \leq N$}{
    Update the posterior probability distribution on $f$ using all available data\\
    Let $\theta_n$ be a maximize point of the $EI$ acquisition function over $\Theta$, where the acquisition function is computed using the current posterior distribution\\
    Observe $y_n = f(\theta_n)$.\\
    Increment $n$\\
}
\end{algorithm}
\subsection{Hyper-parameters optimization}
\label{hpo}
We employ Bayesian optimization in tuning hyper-parameters, where we treat the hyper-parameter search process as a black-box optimization, as shown in Equation~\eqref{eq:black-box}. Specifically, the hyper-parameter set $\theta_{n}$ includes dropout rate, L2 regularization, propagation depth, learning rate, and dimension of hidden units. The utility function $f$ is set to be the average of clients' accuracy. The high-level optimization process is shown in Algorithm~\ref{Bayesian optimization}.
Firstly, we update the posterior probability distribution on $f$ using all the hyper-parameters sets(line 5). Then we calculate the maximize point of the $EI$ acquisition function as the next hyper-parameters groups and observe the value of utility function (line 6 - line 7).
The hyper-parameter tuning time is measured by $T = n*t$, where $t$ denotes the running time of one set of hyper-parameters, $n$ denotes the number of hyper-parameter combinations, and $T$ is the total hyper-parameter tuning time.
Bayesian optimization optimizes the hyper-parameter tuning time by narrowing down the number of combinations $n$ greatly.
\subsection{Putting all together}
To sum up, we conclude the ASFGNN framework in the Algorithm \ref{ASFGNN}. Before the training process, we initialize the hyper-parameters set of clients and server as $\theta_0$. 
First of all, we get the node embeddings $H_{v,t}^i$ for each client $i$ using Algorithm \ref{Separated Learning} (SGNN) with the relevant hyper-parameters set $\theta_n$ (line 5). 
Secondly, we start the training of FGNN model using node embeddings as the inputs and get the average of accuracy ($M_t$) in each training round (line 7). 
The max of $M_t$ is marked as $M(\theta_n)$ (line 9), which is regarded as outputs of black-box. Then, the following input $\theta_{n+1}$ is updated by Bayesian optimization.
Finally, we get the best hyper-parameters set $\theta_N$ and the relevant $M(\theta_N)$.
\begin{algorithm}[ht]
\caption{\textbf{ASFGNN} (Automated separated-federated graph neural network learning)}\label{ASFGNN}
Initialization of hyper-parameters set: $\theta_0$\\
\For{each BO round $n=0,1,...,N$}
{
    \For{each FGNN round $t=1,2,...,T$}
    {
        \For{each client $i\in \mathcal{P}$}
        {
            $H_{v,t}^i \leftarrow SGNN(V^i,E^i,\theta_{n})$\\
        }
        $M_t \leftarrow FGNN(H_{v,t}^i,\theta_n)$\\
    }
    $M(\theta_n) \leftarrow $ MAX($M_t$)\\
    Update $\theta_{n+1} \leftarrow \argmax_{\theta\in\Theta} M(\theta)$\\
}
\Return $\theta_N$,$M(\theta_N)$\\
\end{algorithm} 
\section{Experiment}
\label{experi}
In this section, we empirically compare the performance of our proposed ASFGNN model with the GraphSAGE of Centralized Model~(CM) which is trained using all the data, the traditional Federated Learning model~(FL) and the Separated model~(SP) in which clients can only use their own data without any communications. We aim to answer the following questions.
\begin{itemize}
\item \textbf{Q1:} whether our model (SFGNN) outperforms the CM model, FL model and SP model that is trained on the isolated Non-IID data, including Non-IID label and Non-IID graph?
\item \textbf{Q2:} how the distribution parameter influences the performance of our model?
\item \textbf{Q3:} how the number of clients influences the performance of our model?
\item \textbf{Q4:} how the JS divergence influences the performance of our model? 
\item \textbf{Q5:} how does Bayesian optimization affect the efficiency of parameter tuning comparing with grid search?
\end{itemize}

\subsection{Experimental settings}
\label{setting}
\subsubsection{Framework}
\label{framework}
We construct our experiment on the popular TensorFlow framework \cite{tensorflow2015-whitepaper}. 
All the experiments were performed on a Macbook Pro laptop with 2.3GHz 4-core Intel Core i5 processor. For simplification, we ignore the communication cost and focus on the performance and computation efficiency. 

\subsubsection{Datasets}
\label{datasets}
To test the effectiveness of our proposed model, we choose three benchmark citation datasets, i.e., Cora, Pubmed, and Citeseer. 
For simplification, we assume there are only two clients ($\mathcal{A}$ and $\mathcal{B}$) who split datasets according to label classes and number of neighbours in graph. 
We use $N_1$ and $N_2$ to denote the number of samples in each part. We divide Cora dataset into $Co_1$ and $Co_2$. The first part $Co_1$ has four label categories (theory, reinforcement learning, genetic algorithms, and probabilistic methods) with 1,412 nodes. 
The second part $Co_2$ contains the rest three label categories (possessing neural networks, case based and rule learning labels) with 1,296 nodes. 
We also divide Citeseer and Pubmed datasets into two parts ($Ci_1$ and $Ci_2$, $Pu_1$ and $Pu_2$) in a similar way. We report the data split result in Table \ref{subdataset}.
In order to study the influence of data Non-IID on our method, we use $\alpha $ to denote the label distribution ratio. 
The data of client $\mathcal{A}$ is made up of $\alpha \cdot N_1$ samples from the first part and $(1 - \alpha ) \cdot N_1$ samples from the second part.
Similarly, the data of client $\mathcal{B}$ is made up of $(1 - \alpha ) \cdot N_2$ samples from the first part and $\alpha \cdot N_2$ samples from the second part. 
In other words, the hyper-parameters $\alpha$ implies the non-iid level. We assume that $\alpha$ ranges from 0.5 to 1.0 due to the symmetry.
We use exactly the same dataset split of training, validate, and test following the prior work \cite{kipf2016semi}.
Apparently, our proposal can be applied into the scenario where there are multiple clients. 
\begin{table}
\centering
\caption{Statistic analysis of subsets.}
\label{subdataset}
\begin{tabular}{|c|c|c|c|c|}
  \hline
  Subset & \#Nodes & \#Edges & \#Features & \#Classes \\
  \hline
  \hline
  $Co_1$ & 1,412 & 2,657 & 1,433 & 4 \\
  \hline
  $Co_2$ & 1,296 & 1,961 & 1,433 & 3 \\
  \hline
  $Ci_1$ & 1,507 & 2,024 & 3,703 & 3 \\
  \hline
  $Ci_2$ & 1,805 & 2,005 & 3,703 & 3 \\
  \hline
  $Pu_1$ & 9,791 & 16,585 & 500 & 2 \\
  \hline
  $Pu_2$ & 9,926 & 19,020 & 500 & 2 \\
  \hline
\end{tabular}
\end{table}

\subsubsection{Metrics} 
\label{metrics}
Following the existing work \cite{kipf2016semi}, we use accuracy as the evaluation metric. To compare the performance of different strategies in decentralized scenario, we choose the average of metrics in all clients as the optimization target.

\subsubsection{Hyper-parameters} 
\label{hp}
Following recent research \cite{liu2018geniepath}, we use hyperbolic tangent~(TanH) as the active function of hidden layers and set the max layer of the fully-connected deep neural network in the discrimination model ($L=2$). We tune other hyper-parameters by using Bayesian optimization. The hyper-parameters include dropout rate $d \in \{0.0, 0.5 \}$, L2 regularization $l_2 \in \{0.0, 5e^{-4}, 1e^{-3}, 5e^{-3}, 1e^{-2}\}$, propagation depth $K \in \{1, 2, 3, 4, 5\}$, learning rate $l_r \in \{5e^{-4}, 1e^{-3}, 5e^{-3}, 1e^{-2} \}$, and dimension of hidden units $l \in \{ 64, 128, 256, 512\}$. As clients train the discrimination model federated, the dimension of embeddings should be aligned, which means that all clients have the same hidden units dimension.
The experiment are conducted in a stand-alone PC to simulate the communication in federated learning.
We tune parameters based on the validate dataset and evaluate model performance on the test dataset.

\subsection{Accuracy comparison}
\label{accu}

\subsubsection{Accuracy comparison of different models with Non-IID label}
\label{accunoniid}
To answer the proposed question \textbf{Q1}, we first set the label distribution ratio $\alpha=1.0$, which implies the labels between client $\mathcal{A}$ and client $\mathcal{B}$ are totally different. 
In general, we take advantage of grid search method to find the highest accuracy with the proper hyper-parameters. 
We summarize the comparison results in Table \ref{compare accuracy}, and report the corresponding best hyper-parameters set in Table \ref{compare Hyper-parameters}. 

\begin{table}
\centering
\caption{Performance comparison on three datasets in terms of accuracy.}
\label{compare accuracy}
\begin{tabular}{|c|c|c|c|c|}
\hline
Dataset & CM & FL & SP & SFGNN \\ 
\hline
\hline
Cora & 0.8150 & 0.8833 & 0.9101 & \textbf{0.9264}  \\
\hline
Citeseer & 0.7001 & 0.7500 & 0.7823 & \textbf{0.8055} \\
\hline
Pubmed & 0.7910 & 0.8889 & 0.9174 & \textbf{0.9340}  \\
\hline
Average & 0.7687 & 0.8407 & 0.8699 & \textbf{0.8886}  \\
\hline
\end{tabular}
\end{table}
\begin{table}
\centering
\caption{Hyper-parameters of the SFGNN model and FL model with the best accuracy.}
\label{compare Hyper-parameters}
\begin{tabular}{|c|c|c|c|c|}
\hline
Model & $K$ & $lr$ & $l_2$ & $d$\\ 
\hline
\hline
FL of Cora& 4 & 0.01 & 0.005 & 0.0 \\
\hline
SFGNN of $\mathcal{A}$ on Cora & 4 & 0.01 & 0.005 & 0.5 \\
\hline
SFGNN of $\mathcal{B}$ on Cora & 2 & 0.01 & 0.005 & 0.5 \\
\hline
FL on Citeseer& 4 & 0.005 & 0.005 & 0.0 \\
\hline
SFGNN of $\mathcal{A}$ on Citeseer & 4 & 0.005 & 0.01 & 0.5 \\
\hline
SFGNN of $\mathcal{B}$ on Citeseer & 4 & 0.01 & 0.01 & 0.0 \\
\hline
FL on Pubmed & 5 & 0.005 & 0.001 & 0.5 \\
\hline
SFGNN of $\mathcal{A}$ on Pubmed & 3 & 0.01 & 0.001 & 0.0 \\
\hline
SFGNN of $\mathcal{B}$ on Pubmed & 2 & 0.01 & 0.005 & 0.5 \\
\hline
\end{tabular}
\end{table}


From the Table \ref{compare accuracy}, we can conclude that SFGNN outperforms the other three models in all the datasets. Besides, comparing with the traditional FL model, the improvement of accuracy is about 5.70\% percent in average, which means the SFGNN model is more effective for data Non-IID scenarios. Because our proposed SFGNN generates embeddings separately with individual hyper-parameters and aggregates discrimination layers of clients, the SFGNN model can balance the inference and contributions from samples with different labels. 
From Table \ref{compare accuracy}, we can also find an interesting result. That is, the Centralized GNN Model (CM) achieves the worst performance. This is because clients have absolutely different label classes when the $\alpha = 1.0$, and the models with relatively pure label classes will naturally achieve better performance. When different label classes are combined together in CM, it introduces distractions to the model learning target, which makes CM behave the worst. 

From Table \ref{compare Hyper-parameters}, we can also observe that clients generally have different hyper-parameters to achieve the best accuracy, and these parameters are also different from the hyper-parameters of FL model.
The individual hyper-parameters describe the diversity of Non-IID datasets.

\subsubsection{Accuracy comparison of different models with both Non-IID label and Non-IID graph}
\label{accugraph}
he GNN model benefits a lot from adjacent information, which is different from the traditional neural network model, the distribution of graph data also has an important influence on model accuracy.
As median of edges in Cora dataset is $3.8$, we firstly split the Cora dataset into two sub-datasets, i.e., $Co_3$ and $Co_4$, according to the average edges. The $Co_3$ has the samples with equal or lesser than $3$ edges, while $Co_4$ containts the rest samples. Furthermore, we combine the Non-IID graph data with the Non-IID label data, which means the datasets have different graph and label distributions. 
Similar as the setting in Table \ref{subdataset}, we build a subset of $Co_3$ as $Co_5$, which only has label classes of `theory', `reinforcement learning', `genetic algorithms', and `probabilistic methods', and a subset of $Co_4$ as $Co_6$, which only has labels of `possessing neural networks', `case based' and `rule learning'. 
Similarly, we get the subsets of Citeseer dataset ($Ci_3$, $Ci_4$, $Ci_5$, $Ci_6$).
After the data being preprocessed, client $\mathcal{A}$ owns the $Co_5$ subset and client $\mathcal{B}$ owns the $Co_6$ subset of Cora dataset, and 
client $\mathcal{A}$ owns the $Ci_5$ subset and client $\mathcal{B}$ owns the $Ci_6$ subset of Citeseer dataset.
We train the SFGNN model and FL model respectively and compare their accuracy in Table \ref{compare Non-IID}.
We can conclude that the SFGNN model performs better than FL model when both label and graph are Non-IID, and the improvement percent average increases from 5.70\% to 7.04\%. The experiment results indicate that SFGNN model is more appropriate for the scenarios where both graph and label have different distributions.
\begin{table}
\centering
\caption{Performance comparison on both Non-IID label and Non-IID graph.}
\label{compare Non-IID}
\begin{tabular}{|c|c|c|c|}
\hline
Dataset & FL & SFGNN & Improvement\\ 
\hline
\hline
Cora & 0.7525 & 0.7986 & 6.13\% \\
\hline
Citeseer & 0.7020 & 0.7583 & 8.02\% \\
\hline
Average & 0.7273 & 0.7785 & 7.04\% \\
\hline
\end{tabular}
\end{table}

\subsubsection{Accuracy comparison with different label distribution}
\label{acculabel}
\begin{figure}
  \centering
  \includegraphics[width=0.5\textwidth]{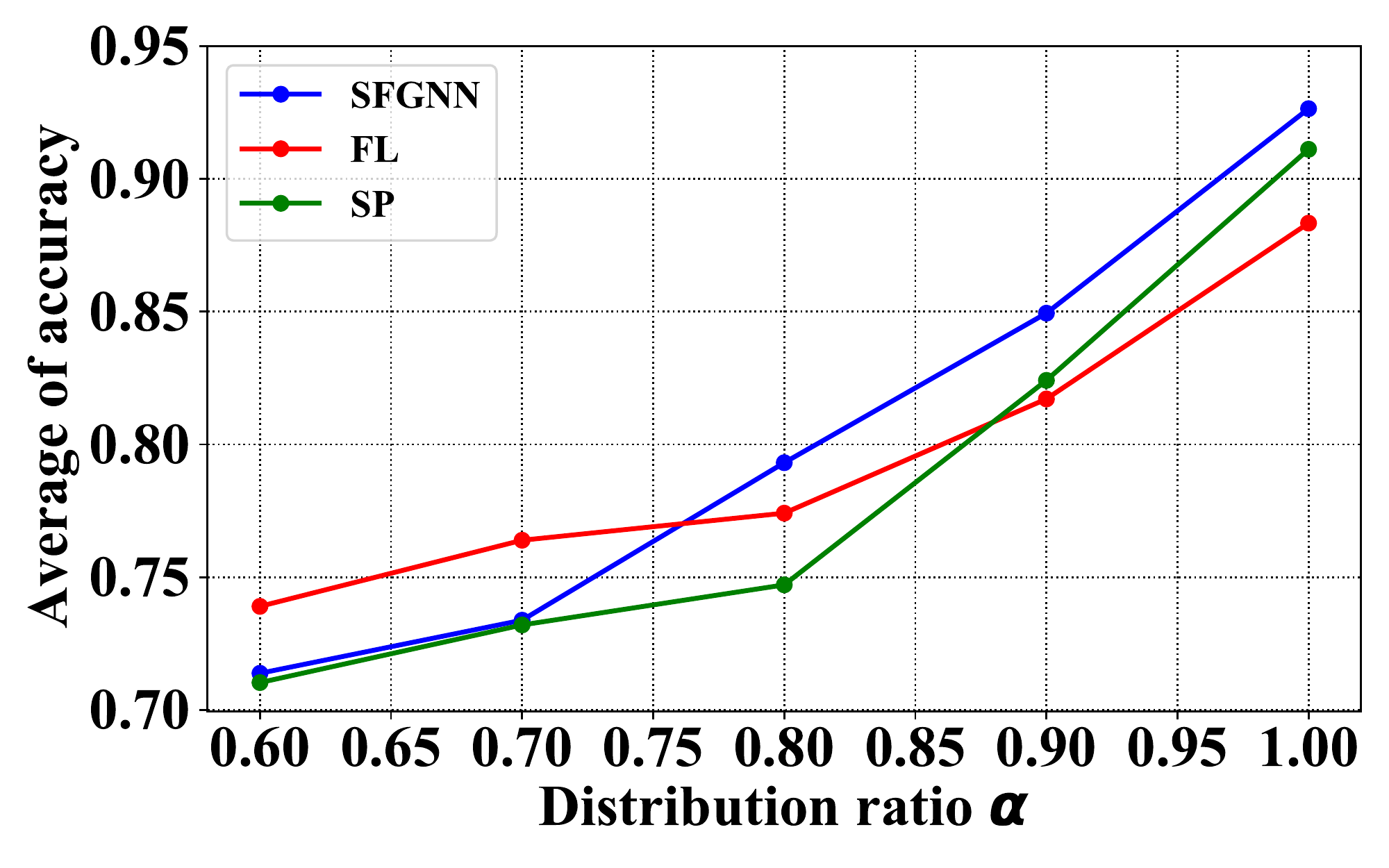}
  \caption{Average accuracy comparison of three models with different $\alpha$.}
  \label{compare_ratio}
\end{figure}

To answer the proposed question \textbf{Q2}, we vary $\alpha$ from 0.6 to 1.0 on Cora dataset and report the accuracy of different models in Fig. \ref{compare_ratio}. We compare our model with the SP model in which client $\mathcal{A}$ and client $\mathcal{B}$ can only use their own data without any communications. 
The bigger ratio means the less similar distributions of clients' datasets. From the results, we can conclude that (1) SFGNN performs better than both SP model and FL model when data distribution is more asymmetrical ($\alpha > 0.75$), which is a quite common situation in real-world applications, (2) FL model is more suitable for training a single global model when all the clients tend to have IID data ($0.5 <= \alpha < 0.75$), and (3) SP model even works better than FL when clients have severely heterogeneous label distributions ($0.88 < \alpha <= 1.0$).

\subsubsection{Accuracy comparison of different clients' number with Non-IID label.}
\label{accclients}
\begin{figure}
  \centering
  \includegraphics[width=0.5\textwidth]{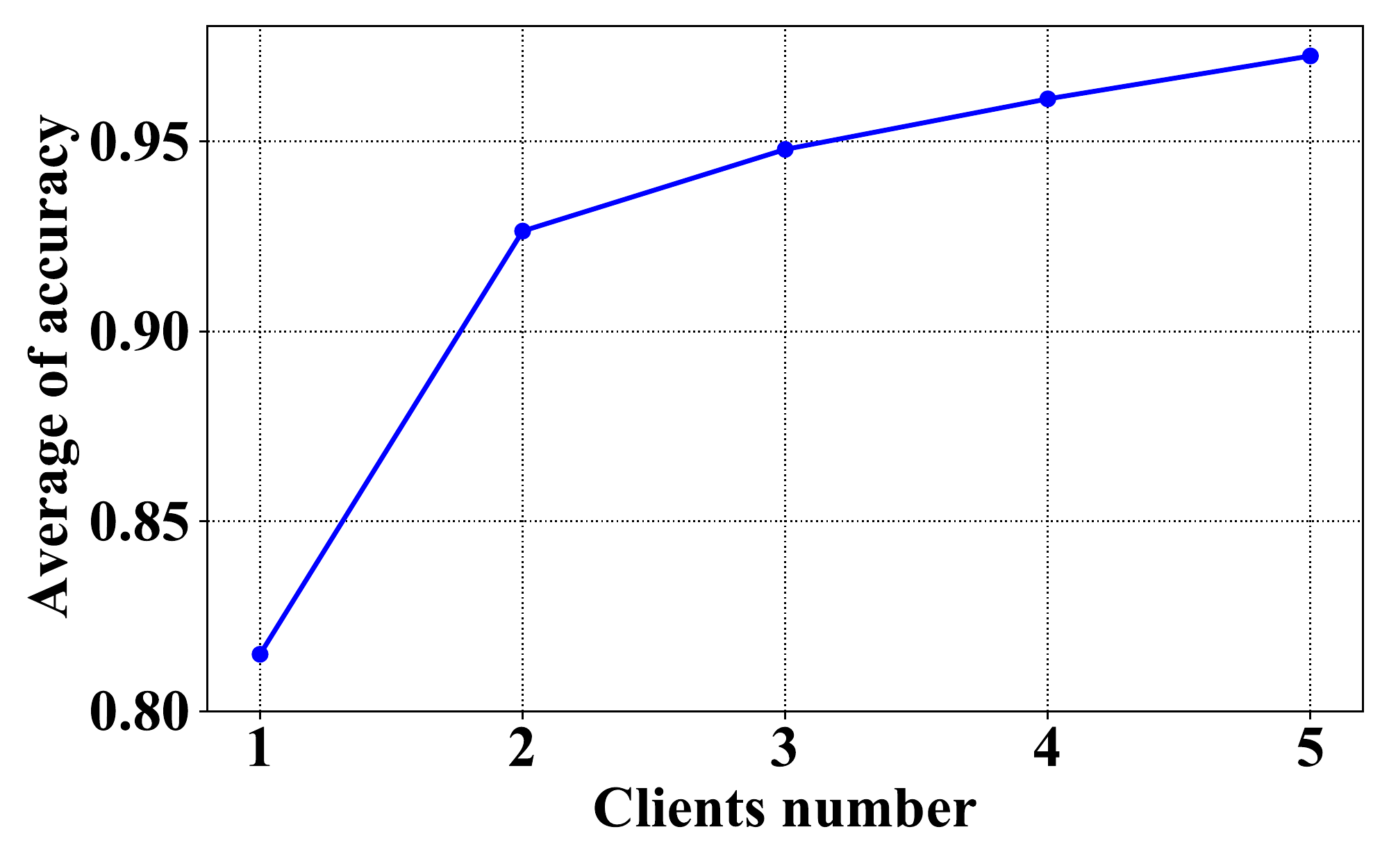}
  \caption{Average accuracy comparison of different clients' number with Non-IID label.}
  \label{compare_number}
\end{figure}

The CM model is trained by the whole dataset, which can be regarded as the ASFGNN model with only one client. To answer the proposed question \textbf{Q3}, we vary the number of clients from 2 to 5 on Cora dataset. The labels of clients' data are different from each other. We report the average accuracy of ASFGNN in Fig. \ref{compare_number}. From it, we can find that the average accuracy of ASFGNN first increases with the number of clients, and then tends to be stable. This is because, when the number of clients first increases, each client has fewer kinds of labels, which makes the Non-IID problem more serious. Therefore, our proposed ASFGNN achieves better performance with the increase of client number. 

\subsubsection{Accuracy comparison of JS divergence}
\label{accujs}
To answer the proposed question \textbf{Q4}, we execute the \textbf{FGNN} model in Algorithm \ref{federated_learning} in an another way. That is, clients update the local discrimination models using global discrimination model directly.
Then we compare the accuracy with different $\alpha$ on Cora dataset, the results are shown in Table \ref{compare_JS}. From it, we find that SFGNN with JS consistently outperforms SFGNN without JS, which shows the effectiveness of the proposed JS method. 
Besides, we also find that the promotion of JS on SFGNN increases with the raise of $\alpha$. 
The contribution of JS method becomes negligible when clients tend to have IID data, e.g., $\alpha=0.6$. 
\begin{table}
\centering
\caption{Performance comparison of JS in SFGNN model.}
\label{compare_JS}
\begin{tabular}{|c|c|c|c|}
\hline
Ratio $\alpha$ & Without JS & With JS & Improvement\\ 
\hline
\hline
1.0 & 0.9081 & 0.9264 & 3.66\% \\
\hline
0.9 & 0.8226 & 0.8494 & 3.26\% \\
\hline
0.8 & 0.7692 & 0.7931 & 3.10\% \\
\hline
0.7 & 0.7120 & 0.7338 & 3.06\% \\
\hline
0.6 & 0.7018 & 0.7138 & 1.72\% \\
\hline
\end{tabular}
\end{table}
\subsection{Efficiency comparison}
\label{efficiency}
To answer the proposed question \textbf{Q4}, we compare the parameters tuning time of grid search and Bayesian optimization on the three datasets. 
We perform the Bayesian optimization of hyper-parameters with the help of the open source framework \textit{SMAC3} \cite{smac2017}. The domains of $l_r, l_2, d$ in \textit{SMAC3} are continuous, and the domains of $K, l$ are discrete with the interval of $1$.
Both grid search method and Bayesian optimization method are implemented under the same computation and communication environment. 
We report the parameter tuning time in Table \ref{compare_time}. Note that both methods achieve comparable accuracy on these three datasets. 
From Table \ref{compare_time}, we can observe that, (1) the Bayesian optimization method greatly reduces the hyper-parameters tuning time on all the three datasets, comparing with the traditional grid search method, and (2) the speedup of Bayesian optimization against grid search becomes higher when dataset gets larger. For example, the speedup on Pubmed dataset is 102.56 while it is 10.10 on Cora dataset. This is because Bayesian optimization reduce the parameter tuning time of grid search by decreasing the parameter search space. In real-world applications, the network bandwidth is always limited between clients and server, and the model training procedure under data isolated setting usually takes much longer time then traditional centralized model training. Therefore, decreasing the parameter search space becomes the key of reducing the tuning time. The results demonstrate that our proposal is good at doing this. 
\begin{table}
\centering
\caption{Training time comparison between BO and grid search on three datasets.}
\label{compare_time}
\begin{tabular}{|c|c|c|c|}
\hline
Datasets & Grid Search & BO & Speedup\\ 
\hline
\hline
Cora & 6.67h & 0.39h & 17.10 \\
\hline
Citeseer & 40.85h & 0.42h &97.26 \\
\hline
Pubmed & 427.67h & 4.17h & 102.56 \\
\hline
\end{tabular}
\end{table}
\section{Conclusion and future network}
In this paper, we proposed a Automated Separated-Federated GNN learning paradigm in the Non-IID isolated scenario. 
We first proposed a separated-federated GNN learning model, which decoupled the training of GNN into two parts: the message passing part was done by clients separately, and the loss computing part was learnt by clients federally. 
To handle the time-consuming problem, we leveraged the Bayesian optimization technique to automatically tune the hyper-parameters of all the clients.
Experiments on real world datasets demonstrated that our model significantly outperformed the federated GNN learning on the isolated Non-IID data.

In the future, we would like to verify our proposal with more existing GNN models. 
We are also interested in deploying our proposal into real-world applications.

\bibliographystyle{unsrt}  

\bibliography{reference}

\end{document}